\documentclass[runningheads]{llncs}
\usepackage{graphicx}
\usepackage{url}
\usepackage{array,ragged2e,booktabs}
\begin{document}
\title{The FRENK Datasets of Socially Unacceptable Discourse in Slovene and English}
\titlerunning{Datasets of socially unacceptable discourse in Slovene and English}
%

%


\author{Nikola Ljube\v{s}i\'{c}\inst{1}\orcidID{0000-0001-7169-9152} \and
Darja Fi\v{s}er\inst{2,1}\orcidID{0000-0002-9956-1689} \and
Toma\v{z} Erjavec\inst{1}\orcidID{0000-0002-1560-4099}}
\authorrunning{N. Ljube\v{s}i\'{c} et al.}
%
\institute{Dept. of Knowledge Technologies, Jo\v{z}ef Stefan Institute, Ljubljana, Slovenia 
\email{\{nikola.ljubesic,tomaz.erjavec\}@ijs.si}
\and
Dept. of Translation, Faculty of Arts, University of Ljubljana, Slovenia
\email{darja.fiser@ff.uni-lj.si}}
\maketitle              
\begin{abstract}
In this paper we present datasets of Facebook comment threads to mainstream media posts in Slovene and English developed inside the Slovene national project FRENK\footnote{The acronym FRENK stands for ``FRENK - Raziskave Elektronske Nespodobne Komunikacije'' (engl. ``Research on Electronic Inappropriate Communication'').} which cover two topics, migrants and LGBT, and are manually annotated for different types of socially unacceptable discourse (SUD). The main advantages of these datasets compared to the existing ones are identical sampling procedures, producing comparable data across languages and an annotation schema that takes into account six types of SUD and five targets at which SUD is directed. We describe the sampling and annotation procedures, and analyze the annotation distributions and inter-annotator agreements. We consider this dataset to be an important milestone in understanding and combating SUD for both languages.

\end{abstract}
\section{Introduction}

With the transformative role of social media in public communication and opinion, there is increased pressure to understand and manage inappropriate on-line content. The research community is by now generally aware of the complexity of the inappropriateness in on-line communication, and there is an overall agreement that annotating real-world datasets with these phenomena, both for understanding the phenomenon via statistical analysis, as well as for \mbox{(semi-)automation} via machine learning is the way forward in combating this major downside of the social media \cite{ljubevsic2018datasets,DBLP:conf/emnlp/PavlopoulosMA17}.

The currently available datasets of inappropriate on-line communication are primarily compiled for English, such as the Twitter dataset annotated for racist and sexist hate speech \cite{DBLP:conf/naacl/WaseemH16}\footnote{\url{https://github.com/ZeerakW/hatespeech}}, the Wikimedia Toxicity Data Set \cite{Wulczyn:2017:EMP:3038912.3052591}\footnote{\url{https://figshare.com/projects/Wikipedia_Talk/16731}}, the Hate Speech Identification dataset 
\cite{DBLP:journals/corr/DavidsonWMW17}\footnote{\url{https://data.world/crowdflower/hate-speech-identification}}, the SFU Opinion and Comment Corpus 
\footnote{\url{https://github.com/sfu-discourse-lab/SOCC}}, and the Offensive Language Identification Dataset (OLID) 
\cite{zampierietal2019}\footnote{\url{https://scholar.harvard.edu/malmasi/olid}}.  Datasets in other languages have recently started to emerge as well, with a German Twitter dataset on the topic of refugees in Germany \cite{DBLP:journals/corr/RossRCCKW17}\footnote{\url{https://github.com/UCSM-DUE/IWG_hatespeech_public}}, a Greek Sport News Comment dataset 
\cite{DBLP:conf/emnlp/PavlopoulosMA17}\footnote{\url{https://straintek.wediacloud.net/static/gazzetta-comments-dataset/gazzetta-comments-dataset.tar.gz}} and two large datasets of Slovene and Croatian online news comments manually moderated by the site administrators \cite{ljubevsic2018datasets}\footnote{\url{http://hdl.handle.net/11356/1201}} \footnote{\url{http://hdl.handle.net/11356/1202}}.

The annotation schemas used in these datasets are very different, ranging from encoding multiple toxicity levels, covert vs. overt aggressiveness, the target of the inappropriateness only etc. The first two pieces of work to take into account both the type of SUD and its target are the annotation schema presented in \cite{fivser2017legal} (which is used in the dataset presented in this paper) and the OLID dataset \cite{zampierietal2019}.

In this paper we present datasets of Facebook posts and comments of mainstream news media from Slovenia and Great Britain, covering the topics of migrants and LGBT. Each comment is annotated with a two-dimensional annotation schema for SUD, covering both the type and the target of SUD. The main contributions of this paper are the following: (1) we offer a selection of Facebook pages aimed at representativeness and comparability for a specific country / language, (2) we apply an identical formalism on comparable data in two languages, making this the first multilingual dataset annotated for SUD we are aware of, (3) we annotate for a very broad phenomenon of SUD, covering most phenomena various datasets cover in isolation, (4) we annotate full discussion (comment) threads, not isolated short utterances, ensuring both that (a) the annotators are as informed of the context as possible while making their decisions (e.g., annotating tweets in isolation, not knowing their context, is a questionable, but regular practice) and (b) that the context of the comment is available either for analyzing the dataset or using the dataset for (semi)automating the identification of SUD, and (5) we perform a first analysis of this rich dataset, observing interesting phenomena both across topics and across languages.


\section{Dataset Construction\label{sec:construction}}

By selecting the Facebook pages of mainstream media that would constitute our dataset, we were aiming at criteria that would make the datasets as comparable as possible across languages.\footnote{While in this paper we describe the annotation results of Slovene and English only, an annotation campaign over Croatian data is already under way and plans exist to annotate Dutch and French data as well.} We opted for the most visited web sites of mainstream media according to the Alexa service\footnote{\url{https://www.alexa.com/topsites/countries}} that have popular Facebook pages. For Slovene, this procedure yielded 24urcom,\footnote{\url{https://www.facebook.com/24urcom}} SiOL.net.Novice\footnote{\url{https://www.facebook.com/SiOL.net.Novice}} and Nova24TV\footnote{\url{https://www.facebook.com/Nova24TV}}, while for English we selected bbcnews,\footnote{\url{https://www.facebook.com/bbcnews}} DailyMail \footnote{\url{https://www.facebook.com/DailyMail}} and theguardian.\footnote{\url{https://www.facebook.com/theguardian}} The only intervention into the list obtained from the Alexa service was the removal of the Slovene public broadcast RTV Slovenija, as its Facebook page RTV.SLOVENIJA is not very active since this medium has a very good in-house solution to news commenting.

Once we harvested all the available posts and comments from these pages via the public Graph API\footnote{\url{https://developers.facebook.com/docs/graph-api/}}\footnote{To use this service from May 2018 onwards, users have to go through a screening process that would quite likely not be successful for harvesting purposes, but our collection was performed in October 2017, before this restrictive change in policy.}, we started the process of identifying posts covering our two topics: migrants and LGBT. Topic identification was performed in the following way: (1) we manually identified a selection of around 100 posts per topic (including a category for covering \emph{other} news, i.e., not migrants and not LGBT) via keywords, (2) we trained a simple word- and character-ngram-based linear SVM classifier on these posts, (3) we classified the whole collection of Facebook posts with that classifier, (4) we manually corrected classifications of 100 random posts per topic (including the \emph{other} category), (5) we added these posts to our dataset of manually-annotated posts and retrained the classifier, and (6) we performed a final classification of the posts. For the final annotation we discarded the posts annotated as \emph{other} and selected the posts annotated with the two remaining topics with the highest confidence score.

With this process which included two rounds of (quick) manual annotation we wanted to produce a training set representative of the two topics, and not of the keywords used in the initial topic identification.

In early stages of the development of the topic identification process, we experimented with (1) merging all Facebook pages together vs. classifying each Facebook page separately and (2) representing each post either through the post text only vs. representing it with the post text as well as the text of the first ten comments. These experiments showed that better results are achieved if (1) all Facebook posts were classified together (having three times the amount of training data, but greater variety of post and comment styles) and (2) each post was represented through the text of the post and the first ten comments (having more text per post at classifier's disposal, but combining the text of the post together with the users' responses, which might be less informative of the topic).


\section{Dataset Annotation\label{sec:annotation}}

For annotating the datasets, we used a two-dimensional annotation schema an early version of which was presented in \cite{fivser2017legal}, covering both the type of potentially socially unacceptable discourse and the target this discourse is aimed at. The annotation was performed in  PyBossa,\footnote{\url{https://pybossa.com}} a web-based crowdsourcing tool.

\subsection{Annotation Schema}

While annotating for the type of SUD, annotators used the decision tree sketched in Figure \ref{fig:type}. This decision tree has six leaves, representing the six types of SUD we discriminate between. The main distinctions regarding the type of SUD are made whether SUD is aimed at the background of a person (e.g., religion, sexual orientation), whether the SUD is aimed at other groups or an individual (vs. just being unacceptable in terms of swearing), and whether there are elements of violence in SUD or not.

\begin{figure}
    \centering
\begin{verbatim}
Is this SUD aimed at someone's background?
    YES: Are there elements of violence?
         YES: background, violence
         NO:  background, offensive speech
    NO:  Is this SUD aimed towards individuals or other groups?
         YES: Are there elements of violence?
              YES: other, threat
              NO:  other, offensive speech
         NO:  Is the speech unacceptable?
              YES: inappropriate speech
              NO:  acceptable speech
\end{verbatim}
\caption{Decision tree used for identifying the type of socially unacceptable discourse}
    \label{fig:type}
\end{figure}

As for the target of SUD, we discriminate between (1) SUD aimed at migrants or LGBT (2) SUD aimed at individuals or groups related to migrants or LGBT (such as NGOs, public bodies etc.), (3) SUD aimed at journalists and media, (4) SUD aimed at other commentators in the discussion thread and (5) SUD aimed at someone else.

We consider this annotation schema to be the most comprehensive schema of SUD-related phenomena applied to this day on any dataset.

\subsection{Annotation Procedure}

\begin{figure}
    \centering
    \includegraphics[width=\textwidth]{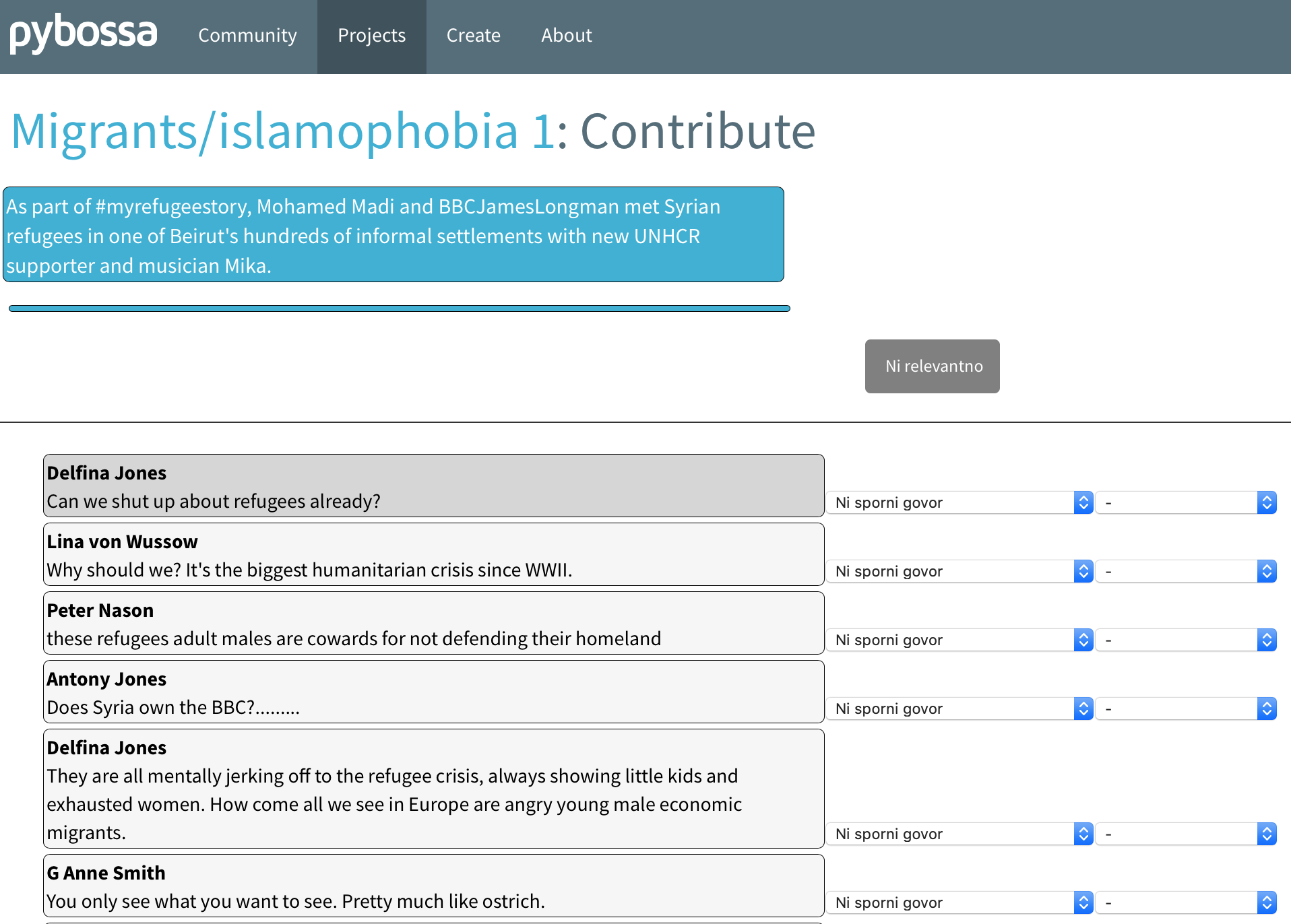}
    \caption{The PyBossa interface used for annotating the datasets.}
    \label{fig:pybossa}
\end{figure}

The annotation campaign was divided into tasks, where each task consisted of the post text (published by the medium) and all the comments (written by their readers) in the discussion thread. For each task the post either had to be annotated as irrelevant (in case of incorrect topic classification presented in Section \ref{sec:construction}), or each comment in the task had to be annotated with the corresponding SUD type and target via two drop-down menus. Going back to tasks that were already submitted was not possible as PyBossa does not offer such an option. Furthermore, discussion threads that were longer than 20 comments were split into multiple tasks as annotating more than 20 comments before submitting a task would easily become too cumbersome for the annotators. The tasks continuing the annotation of a discussion thread also contained the five last comments from the previous task to give the annotators some discussion context. Whole discussion threads were annotated regardless of their length as such an approach was considered to be the least problematic regarding sampling decisions (e.g., what part of the discussion thread to annotate). 
An example screenshot of the annotation interface is depicted in Figure \ref{fig:pybossa}.

The annotators performing the annotation were master students from the Faculty of Arts and the Faculty of Social Sciences from the University of Ljubljana. They all attended an initial half-day annotator training which was followed by an initial annotation round where all of the 32 annotators annotated the same data and discussions about the problematic cases were held via a mailing list. After the initial training, the annotators were randomly assigned to one of the two topics (migrants or LGBT). The same split by topic was then retained throughout the entire annotation campaign. In the first annotation campaign, Slovene data were annotated by approximately eight annotators per comment. The overall decision to gather around eight annotations per comment was based on the awareness of the complexity of the annotation schema and the phenomenon itself, as well as initial calculations of inter-annotator (dis)agreement.
After the annotation campaign of Slovene data, roughly half of the annotators signed up for the annotation of the English data, depending both on their previous performance and their self-assessed skill of the English language. To keep the number of annotations per comment similar to Slovene data, the English annotation campaign took twice as long as the Slovene annotation campaign.

After finishing each PyBossa project, the annotations were downloaded from the PyBossa instance and analyzed, measuring the agreement inside each comment as the entropy of the annotation distribution. Comments with the highest entropy were then manually analyzed by an expert (social scientist working at the national centre for reporting hate speech) in order to communicate the disagreement issues via a mailing list with all the annotators, regardless of the topic they covered in order to fine-tune annotation guidelines and avoid similar issues in future annotation rounds.

\section{Dataset Analysis\label{sec:analysis}}

In this section we perform various analyses of the manually annotated datasets. We start with the description of the size of the dataset, continue with the analysis of the distribution of annotations and the inter-annotator agreement of the two dimensions of annotation, and wrap up with an analysis of the distribution of SUD among the users producing the comments.

\subsection{Size of the Datasets}

We first report on the overall size of the datasets in terms of the number of posts, number of comments and the number of annotator responses in Table \ref{tab:size}. The results show that the length of discussion threads is similar across topics and languages, with the LGBT topic in Slovene being the only strong deviation, where the average discussion thread length is four to five times shorter than the rest of the topics and languages.

\begin{table}
\caption{Size of the datasets by language and topic.}\label{tab:size}
\begin{tabular}{|l|>{\RaggedLeft}p{1.8cm}>{\RaggedLeft}p{1.8cm}|>{\RaggedLeft}p{1.8cm}  >{\RaggedLeft}p{1.8cm}|}
\hline
& \multicolumn{2}{c|}{Slovene} & \multicolumn{2}{c|}{English} \\
& migrants & LGBT & migrants & LGBT \\
\hline
Number of posts & 30 & 93 & 16 & 14 \\
Number of comments & 6545 & 4571 & 5855 & 5906 \\
Average thread length & 218 & 49 & 366 & 422 \\
Number of annotations & 56211 & 41433 & 44969 & 49978 \\
Average number of annotations & 8.59 & 9.06 & 7.68 & 8.46 \\
\hline
\end{tabular}
\end{table}

Regarding the distribution of the sources (Facebook pages) of the posts that were annotated, they are all rather evenly distributed, except for the topic of LBGT in the Slovene dataset, where the number of posts from the Nova24TV is significantly higher than is the case for the other two sources. The Nova24TV source, is associated with the extreme right of the political spectrum, discussing LGBT issues, especially the Slovene referendum on same-sex marriage, more fervently than the remaining sources.

\subsection{Annotation Distribution}
\label{sec:annotation_distribution}

In this subsection we perform an analysis of the distribution of all the annotations by topic and language. Given that there is a large number of possible combinations regarding the type and target of the SUD that we annotate for, Table \ref{tab:annotation} lists only the five most frequently assigned annotation combinations in both languages, treating the two topics as portions of the annotation probability distributions of all the annotation combinations.

\begin{table}
\caption{Probability distribution of the most frequent annotations by language and topic.}\label{tab:annotation}
\begin{tabular}{|p{3.5cm}|p{2.5cm}|>{\RaggedLeft}p{1.35cm} >{\RaggedLeft}p{1.35cm}|>{\RaggedLeft}p{1.35cm} >{\RaggedLeft}p{1.35cm}|}
\hline
& & \multicolumn{2}{c|}{Slovene} & \multicolumn{2}{c|}{English} \\
type & target & migrants & LGBT & migrants & LGBT \\
\hline
Acceptable & No target & 0.42 & 0.54 & 0.50 & 0.67 \\
Background, violence & Target & 0.07 & 0.02 & 0.02 & 0.00 \\
Background, offensive & Target & 0.23 & 0.17 & 0.21 & 0.12 \\
Other, offensive & Commenter & 0.08 & 0.08 & 0.06 & 0.11 \\
Other, offensive & Related to & 0.04 & 0.02 & 0.02 & 0.01 \\
\hline
\end{tabular}
\end{table}

The results show that about half of the annotated comments are socially acceptable, which is a surprisingly low result. It is true that the topics where chosen with the expectation of high SUD occurrence, but we consider the amount of SUD in roughly half of the comments to be rather striking, especially given the fact that some of the most unacceptable comments might have been removed from Facebook prior to our data harvesting.

In Slovene there seems to be less socially acceptable content on these topics, which is observed more frequently on the topic of LGBT than migrants in both languages. The difference between the two languages is especially visible on the LGBT topic where there is an absolute difference of 13\% between Slovene and English, pointing towards greater acceptance of LGBT in the British society.\footnote{As always, these results have to be taken with caution and not as final, as other factors might have produced this difference, such as (1) the fact that in Slovenia the referendum regarding same-sex marriages was carried out during the period these Facebook posts cover and (2) the fact that most of the LGBT-related content comes from Nova24TV, which is, as already mentioned, a medium on the right side of the political spectrum. The latter has proven to have an impact as this source has socially acceptable comments in 42\% of cases, while the other two have 57\% and and 62\% of non-SUD comments on this topic. Both other sources still have, however, a higher percentage of SUD comments than the English average.}

The four most frequently occurring category combinations of SUD in both languages and topics are, in decreasing order: (1) SUD directed at migrants or LGBT people, being offensive, (2) SUD directed at other commenters, being offensive (3) SUD directed at migrants or LGBT people, inciting violence and (4) SUD directed at people or organisations related to migrants and LGBT people, being offensive.

The most worrisome result among these is that the most problematic category, SUD inciting violence towards migrants or LGBT people, is present in the top five categories, covering between 7 and below 1 percent of the annotations, depending on the language and topic. It follows the overall trend of observing higher numbers for Slovene compared to English, and for migrants compared to LGBT. The overall good side of these results is that in the English dataset on the topic of LGBT the percentage of such annotations is below 1\%, but the percentage of such annotations in Slovene related to migrants is at very high 7\%.

Regarding the SUD directed at fellow commenters, in Slovene it is similarly present on both topics, while in English data it is similarly frequent on the topic of migrants, but almost as twice as high on the topic of LGBT. This suggests that the SUD generated on the LGBT topic in English data is often generated due to disagreement with the remaining commenters, not directed at LGBT people directly. This result also explains the lower percentage of more problematic types of SUD for this language and topic. It seems that British people (or, to be more precise, commenters on British online media) are not directly intolerant towards LGBT people \emph{per se}, but are intolerant towards related issues, such as same-sex marriage, same-sex partners adopting children etc.

\subsection{Inter-annotator Agreement}

To measure the complexity of the annotation task as well as the level up to which we should trust single annotations performed in this dataset, we calculated the Krippendorff's $\alpha$ inter-annotator agreement score. This score is suitable for annotation campaigns with more than two annotators, as well as spotty annotations, taking also into account agreement by chance \cite{krippendorff04}. Given that the type of SUD can be considered both a nominal as well as an ordinal variable, we perform calculations of inter-annotator agreement for both cases. For the target of the SUD, no order can be established, therefore we consider this variable to be nominal only.

\begin{table}
\caption{Krippendorff's $\alpha$ inter-annotator agreement by language and topic.}\label{tab:annotation}
\begin{tabular}{|p{2cm}|p{2cm}|p{2cm}p{2cm}|p{2cm}p{2cm}|}
\hline
& & \multicolumn{2}{c|}{Slovene} & \multicolumn{2}{c|}{English} \\
Dimension & Variable type & migrants & LGBT & migrants & LGBT \\
\hline
type & nominal & 0.499 & 0.500 & 0.436 & 0.357 \\
type & ordinal & 0.595 & 0.590 & 0.516 & 0.444 \\
target & nominal & 0.528 & 0.507 & 0.445 & 0.380 \\
\hline
\end{tabular}
\end{table}

The measured agreement between the annotators is considered low by social science standards, i.e., not good enough to draw even tentative conclusions. This is why we have decided to gather multiple annotations from the beginning, with the goal of combining them into single high-quality annotations. Some initial experiments on collecting professional annotations for a subset of Slovene comments and comparing those to the mode of all non-professional annotations, i.e., the most frequent annotation, show that collecting multiple non-professional annotations is a reasonable approach, with the Krippendorff's $\alpha$ for the type of SUD as a nominal variable being 0.731, for the type of SUD as an interval variable 0.795, and for the target of SUD as a nominal variable the Krippendorff's $\alpha$ being 0.733. Thereby, all the agreements are above 0.66, which is the lower threshold for useful annotations by social science standards. Obtaining high-quality annotations from the non-professional annotations is, however, not the focus of this work and will be performed in more detail in future experiments.

Regarding the difference in the observed agreements, it tends to be higher on Slovene than on English which can be due to the fact that the annotators were native speakers of Slovene, living in Slovenia, with better linguistic but also contextual knowledge. As expected, if we consider the type of SUD to be an ordinal, and not a nominal variable, the agreement increases, which shows that the type of SUD does indeed have properties of an ordinal variable. Annotating the type and target of SUD seem to be similarly complex tasks for the annotators. The only deviation in agreement within a language seems to be on English data, where we observe a ten-point lower result for the LGBT topic than for migrants. A tentative hypothesis for such a result is the distribution of the annotations, with the three most frequent annotation combinations making 91\% of the annotations, while in the remaining three cases these cover between 75 and 81 \% of the annotations only. Thereby the agreement by chance in the LGBT topic on English data is higher, making the Krippendorff's $\alpha$ with the same observed agreement lower than for the remaining cases.

As already briefly discussed, future work is planned in increasing the quality of the annotations by exploiting the fact that we have around 8 annotations available per comment which can be used to calculate some statistic, e.g., the mode (most frequent case) of the annotations, which has preliminarly showed to significantly improve the quality of annotations, or to learn to extract the optimal annotation from the multiple annotations by using a small subset of expert annotations in a supervised machine learning setting.

\subsection{Distribution of SUD among Users}

While this dataset enables a myriad of analyses of the occurrence of various forms of SUD due to metadata richness, we have chosen one of the most pressing questions for the first analysis of this dataset: Who produces most of SUD? Is this a small number of users, or is SUD generated in a similar intensity by most users?

\begin{figure}
    \centering
    \includegraphics[width=0.48\textwidth]{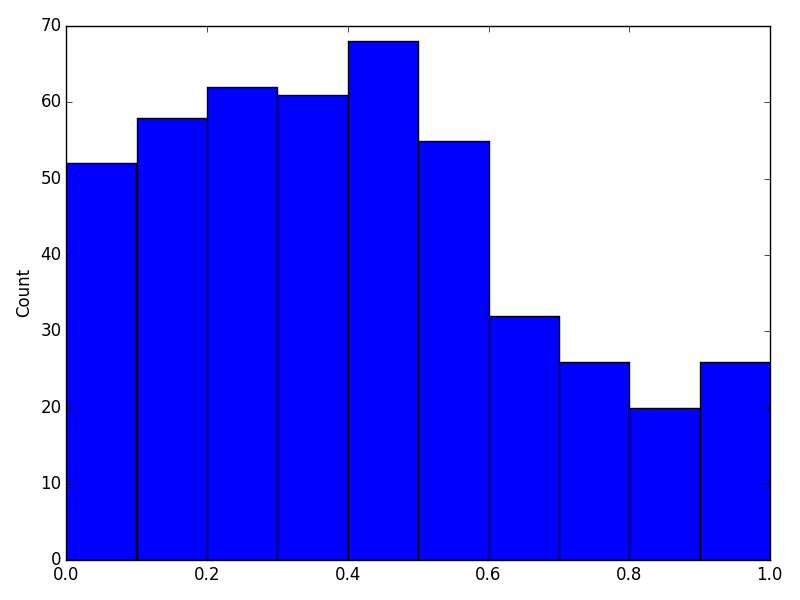}
    \includegraphics[width=0.48\textwidth]{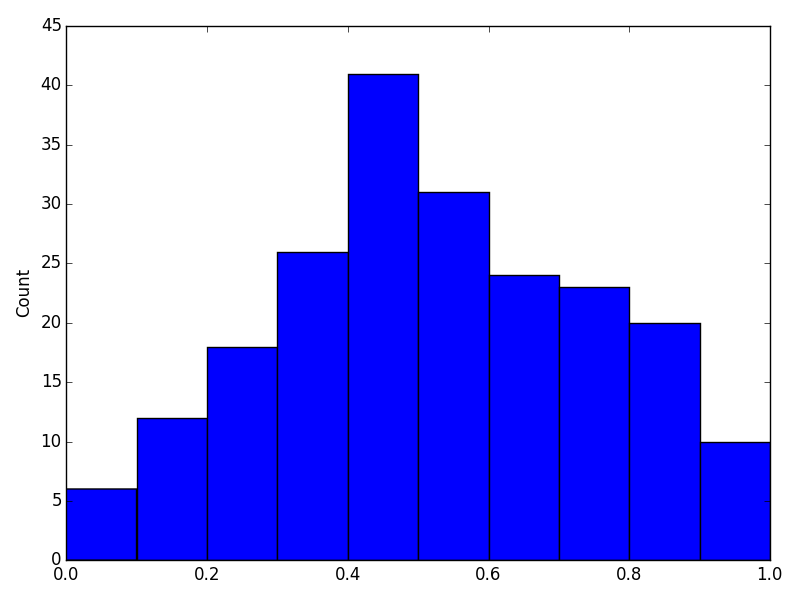}
        \includegraphics[width=0.48\textwidth]{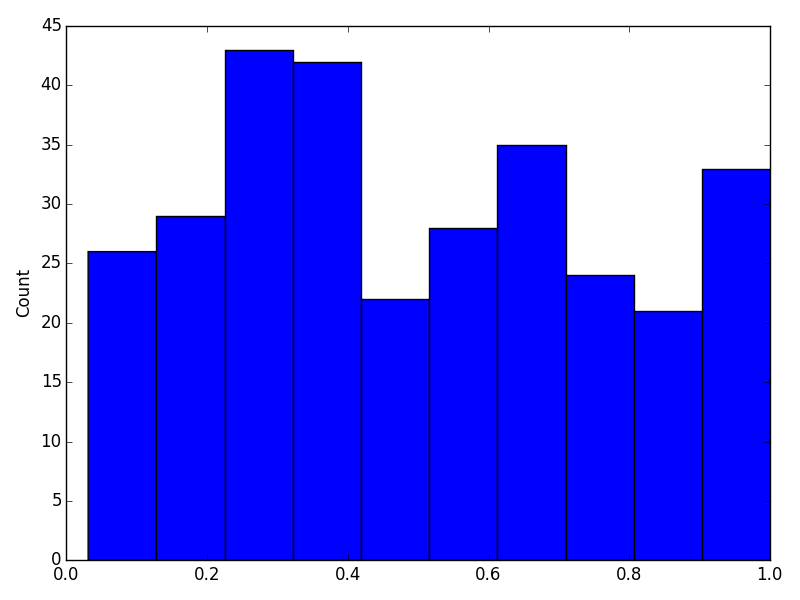}
    \includegraphics[width=0.48\textwidth]{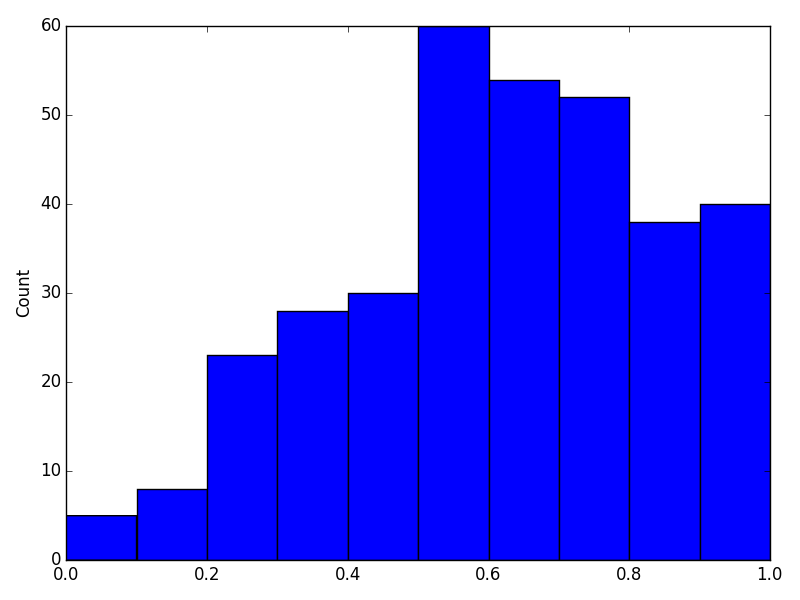}
    \caption{Distribution of percentage of non-SUD comments among users of Slovene (left) and English (right) datasets on topics of migrants (up) and LGBT (down).}
    \label{fig:users}
\end{figure}

We operationalise the amount of SUD produced by specific users as the percentage of non-SUD comments for each user who has posted 3 or more comments in a specific topic and language. We present the results via histograms of percentages of non-SUD comments per user presented in Figure \ref{fig:users}, where the left histograms show the results for Slovene, right for English, upper for the topic of migrants and lower for the topic of LGBT. 

When reading these plots, one has to bear in mind the percentage of non-SUD annotations from Section \ref{sec:annotation_distribution}, which was the lowest for the topic of migrants in Slovene (upper left histogram) and highest for the topic of LGBT in English (lower right histogram). What these plots are primarily informative for is their shape, i.e., whether they are (1) bell-shaped, which points to the conclusion that the majority of users produces a medium amount of SUD, while just a smaller portion produces very little or a lot of SUD, (2) U-shaped, which means that there are users producing either large or small quantities of SUD or (3) rather evenly-distributed, which indicates that users produce various quantities of SUD. The regularity, which can be observed in Figure \ref{fig:users} is that in the English dataset, regardless of the amount of SUD produced, the distributions are rather bell-shaped, meaning that most users are average producers of SUD, while in the Slovene dataset the distributions are rather even, suggesting that users are producing various amounts of SUD.

The results of this analysis point towards the conclusion that English users are more mainstream-oriented regarding the production of SUD, while among the Slovene users there are both extreme, medium and non-producers of SUD. A possible explanation would be for the British media to show a more mature community of commenters than is the case for the Slovene media. Another explanation for the differing distributions between Slovene and English might be a more responsible policy of British media, removing regularly a significant portion of unacceptable content. This phenomenon will have to be investigated in more detail in future work.

\section{Conclusion}

In this paper we have presented a dataset of Facebook comments on posts from mainstream media in Slovenia and Great Britain covering the topics of migrants and LGBT, which were manually annotated with the type and target of socially unacceptable discourse. We have described the data sampling, topic identification and manual annotation of the dataset.

We have performed an initial analysis of the manually annotated dataset, showing that more SUD is produced in Slovene than in English media, and that more SUD is produced on the topic of migrants than LGBT.

Our analysis of the inter-annotator agreement shows that both type and target are similarly complex for the annotators, with a medium agreement below the expected quality in social science. Initial experiments on comparing professional annotations with the modes (most frequent annotations) of the non-professional annotations show to improve agreement significantly, moving it to the area of useful annotations for social sciences. The fact that we have collected around eight annotations per comment will be exploited further in future work on extracting annotations of highest quality possible.

Finally, we have performed an initial analysis of the distribution of SUD among different users, showing that there is a trend of a similar number of users who produce large, medium and low quantities of SUD in Slovene, while the English users tend to produce mostly an average amount of SUD, with just small numbers of users on the extremes of the (non)production of SUD.

Future work on this dataset will focus primarily on the two envisaged usages of the dataset: (1) further analysis of the phenomenon of SUD and (2) \mbox{(semi-)automation} of SUD identification.

\section*{Acknowledgement}

The work described in this paper was funded by the Slovenian Research Agency within the national basic research project ``Resources, methods and tools for the understanding, identification and classification of various forms of socially unacceptable discourse in the information society'' (J7-8280, 2017- 2020) and the Slovenian-Flemish bilateral basic research project ``Linguistic landscape of hate speech on social media'' (N06-0099, 2019 – 2023).

%
%
%
 \bibliographystyle{splncs04}
 \bibliography{mybibliography}
\end{document}